\documentclass[10pt,twocolumn,letterpaper]{article}

\usepackage{iccv}
\usepackage{times}
\usepackage{epsfig}
\usepackage{graphicx}
\usepackage{amsmath}
\usepackage{amssymb}
\usepackage{mathtools}
\usepackage{soul}
\usepackage[table,x11names]{xcolor}
\usepackage{subcaption}

\newcommand*{\modelname}{Multiscale Video Pretraining\xspace}
\newcommand*{\modelabb}{MVP\xspace}

\usepackage[breaklinks=true,bookmarks=false]{hyperref}

\iccvfinalcopy 

\def\iccvPaperID{7057} 

\ificcvfinal\fi

\begin{document}

\title{Multiscale Video Pretraining for Long-Term Activity Forecasting}

\author{Reuben Tan$^{1}$ \ \ \ \ Matthias De Lange$^{2}$ \ \ \ \ Michael Iuzzolino $^{3}$ \ \ \ \ Bryan A. Plummer$^{1}$ \ \ \ \  \\  Kate Saenko$^{1,3}$ \ \ \ \
Karl Ridgeway$^{3}$ \ \ \ \ Lorenzo Torresani$^{3}$ \\
$^{1}$Boston University, $^{2}$KU Leuven, $^{3}$Meta\\
{\tt \small \{rxtan,bplum, saenko\}@bu.edu}, {\tt \small \{matthias.delange\}@kuleuven.be} \\ {\tt \small \{mliuzzolino, karl.ridgeway, torresani\}@meta.com} \\
} 

\maketitle

\maketitle
\ificcvfinal\thispagestyle{empty}\fi

\begin{abstract}
       Long-term activity forecasting is an especially challenging research problem because it requires understanding the temporal relationships between observed actions, as well as the variability and complexity of human activities. Despite relying on strong supervision via expensive human annotations, state-of-the-art forecasting approaches often generalize poorly to unseen data. To alleviate this issue, we propose \modelname (\modelabb), a novel self-supervised pretraining approach that learns robust representations for forecasting by learning to predict contextualized representations of future video clips over multiple timescales. MVP is based on our observation that actions in videos have a multiscale nature, where atomic actions typically occur at a short timescale and more complex actions may span longer timescales. We compare MVP to state-of-the-art self-supervised video learning approaches on downstream long-term forecasting tasks including long-term action anticipation and video summary prediction. Our comprehensive experiments across the Ego4D and Epic-Kitchens-55/100 datasets demonstrate that MVP outperforms state-of-the-art methods by significant margins. Notably, MVP obtains a relative performance gain of over 20\% accuracy in video summary forecasting over existing methods. 
\end{abstract}

\section{Introduction}
\begin{figure}[t]
    \centering
    \includegraphics[width=0.5\textwidth]{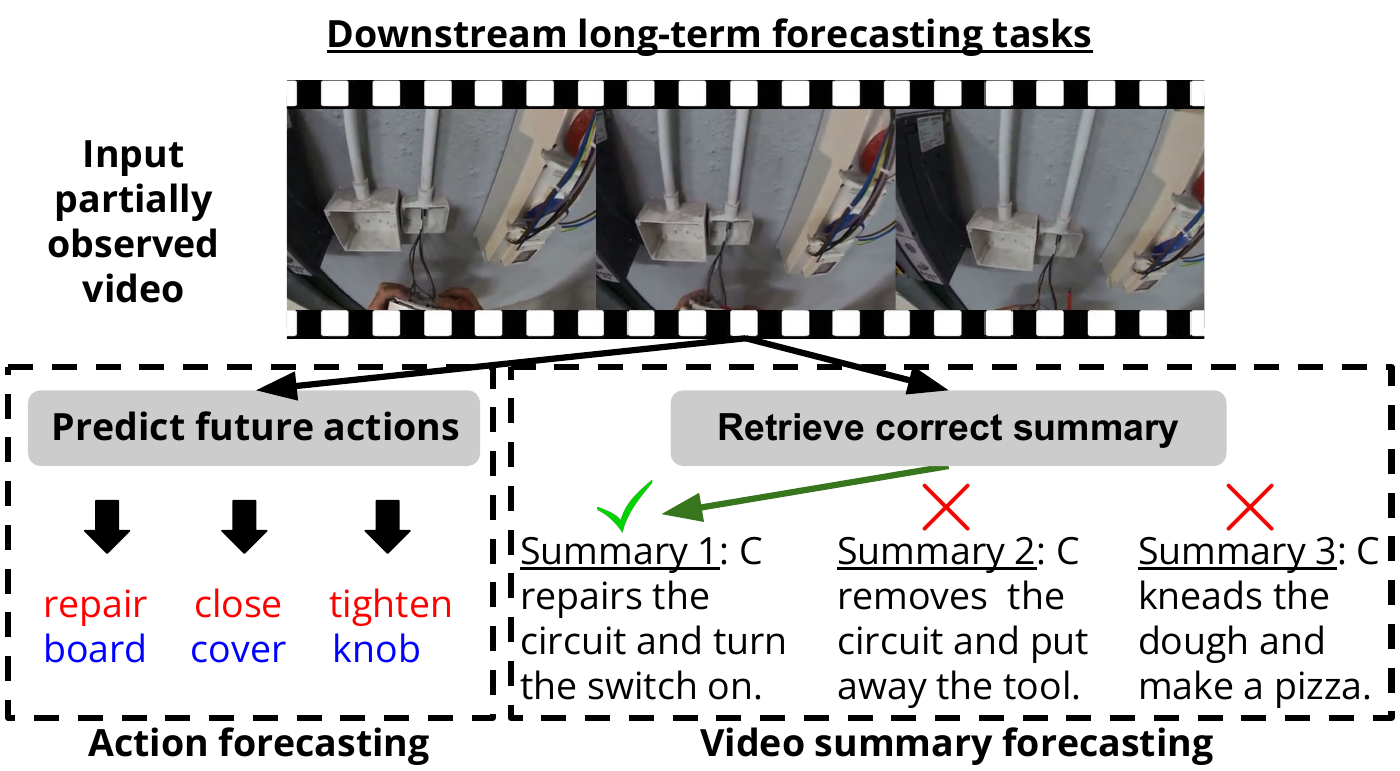}
    \caption{\textbf{Long-term activity forecasting tasks.} We pretrain a video model and transfer its learnt representations to long-term action and video summary forecasting. `C' denotes the camera-wearer in the summaries.}
    \vspace{-10pt}
    \label{fig:forecasting_motiv}
\end{figure}

Long-term forecasting of human activities (illustrated in Figure~\ref{fig:forecasting_motiv}) is a key capability that is crucial for developing intelligent and collaborative machines. Machines that reason about future actions given some observations are better able to plan their own behavior accordingly and interact more effectively with other agents in dynamic environments. However, forecasting future actions is inherently challenging. To begin, the model has to understand the current state of the environment under partial observability. More importantly, the non-deterministic nature of the future compounds the difficulty of having to infer the relationships between actions and objects observed over time and also predict how these relationships will evolve in the future. State-of-the-art long-term forecasting methods (\eg, \cite{gong2022future,grauman2022ego4d}) have focused on learning more effective functions for modeling long-term temporal dynamics in videos by leveraging fully attentional models \cite{vaswani2017attention}, but still rely on pretrained visual representations that are learnt using the standard objective developed for action recognition. However, this objective often encourages a video model to only understand short-term dependencies in a short clip instead of capturing long-term interactions and dynamics of the video. This may limit the generalizability of the pretrained visual representations to long-term forecasting tasks. Despite relying on strong training supervision from human-annotated action labels, the above-mentioned approaches still generalize poorly to unseen data \cite{gong2022future}, which lends support to our theory.


\begin{figure}[t!]
    \centering
    \includegraphics[width=0.5\textwidth]{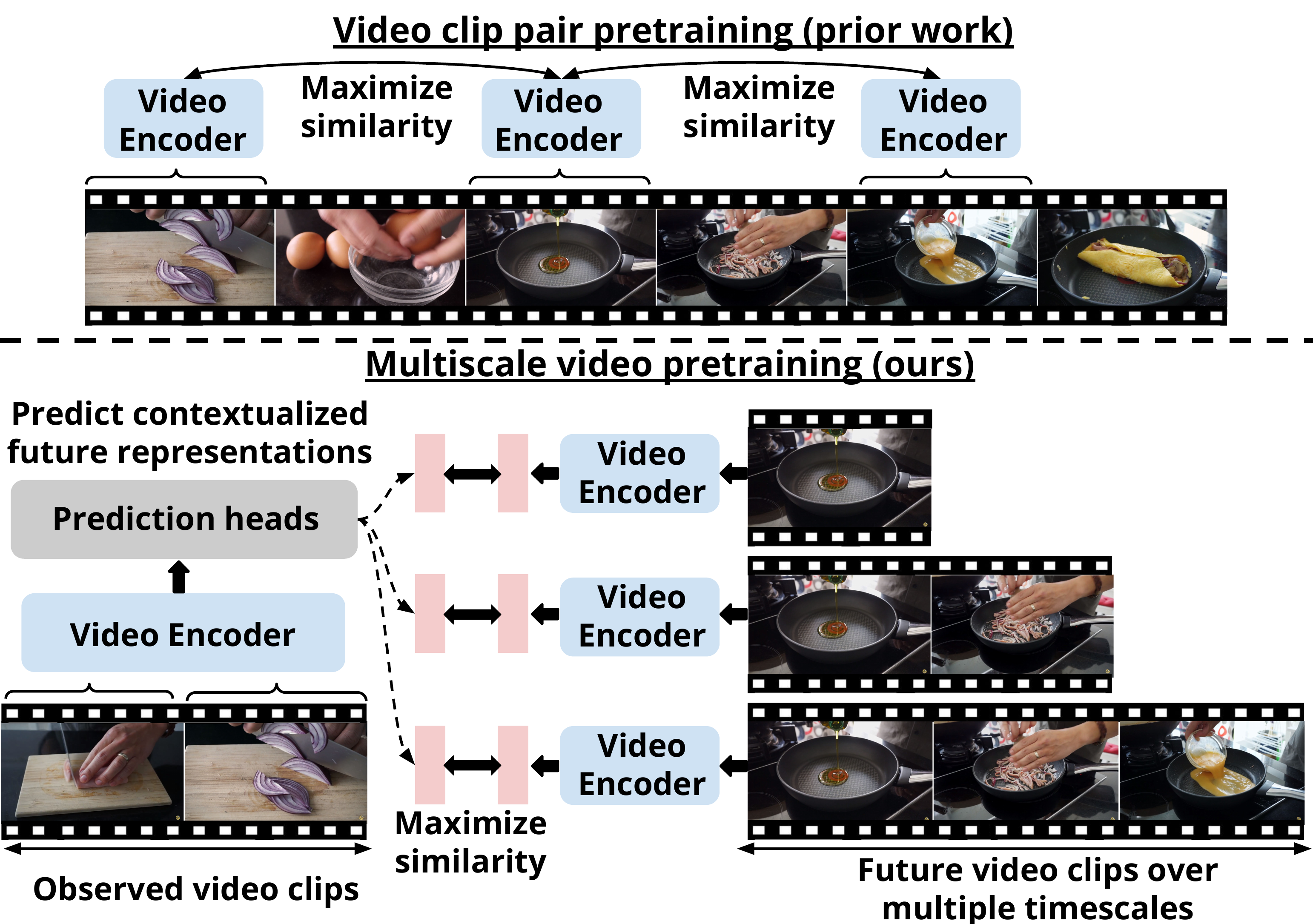}
    \caption{\textbf{\modelname (\modelabb).} In contrast to prior self-supervised methods \cite{qian2021spatiotemporal,feichtenhofer2021large} that maximize the similarity between representations of clips from the same video, \modelabb trains a model to predict future contextual information over different time scales, helping it to generalize better to long-term forecasting tasks.}
    \label{fig:motiv}
    \vspace{-10pt}
\end{figure}

To improve pretraining for long-term forecasting, we first make the observation that videos generally have a \emph{multiscale} nature, where actions can be decomposed into sub-actions that occur at different levels of granularity. Consider Figure~\ref{fig:motiv} that depicts a video of someone preparing a meal. At the highest level of abstraction, the complex action of making an omelette comprises multiple actions, which generally occur at shorter timescales, such as cracking eggs and adding oil. We hypothesize that learning to understand this structure may be crucial for inferring the underlying goals and intentions of the agents involved, thus facilitating more accurate predictions of their subsequent actions. We endeavor to encode the multiscale nature of actions into the learnt video representations in a self-supervised manner during pretraining, which will generalize more effectively to downstream long-term forecasting tasks.

To this end, we introduce a novel \modelname (\modelabb) approach (illustrated in Figure~\ref{fig:motiv}), which encourages a video model to learn to predict contextualized representations of future video clips that have been aggregated over different timescales using information from a partially observed video sequence. MVP draws inspiration from the required capability in long-term forecasting tasks, which necessitates being able to reason about the spatial and temporal structures of observed actions and predict future events that occur over multiple scales and temporal resolutions. During pretraining, \modelabb learns to infer the knowledge from an observed clip sequence that is required to predict the contextual information contained in future clips. 

Given the lack of ground-truth labels in our self-supervised formulation, we generate prediction targets by computing contextualized representations of future video clips. This key aspect of MVP distinguishes it from the state-of-the-art video pretraining objective of maximizing the similarity between representations of different clips sampled from the same video \cite{qian2021spatiotemporal,feichtenhofer2021large} (Figure~\ref{fig:motiv} top). Feichtenhofer \etal \cite{feichtenhofer2021large} demonstrate that the latter objective encourages different clips of the same video to have similar representations over the spatiotemporal dimensions. While learning clip-invariant video representations may be beneficial to the task of short-term action recognition, they do not encode the high-level semantic structure of the observed video. In contrast, the MVP learning objective trains the video model to extrapolate future information at multiple scales from an observed video sequence. By recognizing the relations between different actions in long videos at different levels of granularity, the video model can better understand the underlying structure of videos and make more accurate predictions about what will happen next.


We evaluate the effectiveness of MVP by transferring its pretrained representations to downstream long-term forecasting tasks including order-agnostic and specific action forecasting (Figure~\ref{fig:forecasting_motiv}). Furthermore, we also introduce the novel multimodal task of video summary forecasting, where the goal is to retrieve the corresponding textual summary of the observed and future activities from a set of distractors. MVP significantly outperforms state-of-the-art video pretraining approaches across the Ego4D and Epic-Kitchens-55/100 datasets. More importantly, we extract key insights on the contributions of different aspects of \modelabb through an extensive ablation study that we hope will be beneficial to future work on learning multiscale video representations.

\section{Related work}
\noindent\textbf{Self-supervised video pretraining.} Self-supervised video pretraining \cite{feichtenhofer2021large,han2019video,wang2022long} has been demonstrated to be beneficial for improving performance on downstream tasks such as activity recognition \cite{fan2021multiscale,feichtenhofer2019slowfast,feichtenhofer2021large,han2019video,han2020memory,recasens2021broaden,wang2022long}, video object segmentation \cite{jabri2020space}, early action prediction \cite{shao2020finegym} and unintentional action detection \cite{han2019video,han2020memory} on target datasets including Kinetics-400/600 \cite{carreira2018short,carreira2019short,kay2017kinetics}, HMDB-51 \cite{kuehne2011hmdb} and UCF101 \cite{soomro2012ucf101}. Inspired by image-based self-supervised pretraining objectives \cite{chen2020simple, chen2020improved, chen2021exploring}, state-of-the-art video approaches \cite{feichtenhofer2021large,qian2021spatiotemporal,wang2022long,yuan2022contextualized} often use a similar learning objective of maximizing the similarity between representations of two clips sampled from the same video. The Contrastive Video Representation Learning (CVRL) \cite{qian2021spatiotemporal} approach also demonstrates that the applied transformations have to be consistent across all frames for optimal performance. 

Feichtenhofer \etal \cite{feichtenhofer2021large} also demonstrate that this objective of learning video clip-invariant representions can be extended beyond pairs of clips, which further improves the robustness of the learnt representations to the downstream task of action recognition. Additionally, the Contrastive Predictive Coding (CPC) \cite{oord2018representation} and Dense Predictive Coding (DPC) \cite{han2019video} approaches are also similar in spirit, where their learning objectives are to predict coarse clip-level and fine-grained spatiotemporal region representations of future clips given an observed sequence of clips for context, respectively. Han \etal \cite{han2020memory} further build on this by introducing a memory bank of learnable vectors to account for the non-deterministic nature of predicting the future. However, in contrast to our MVP approach, the aforementioned approaches learn to predict the information in the future clips that directly follow after the observed sequence. More importantly, they only predict the base representations of future video clips instead of their \emph{contextualized} representations, where their information has been aggregated over all preceding future clips in a causal manner. 

Additionally, BraVe \cite{recasens2021broaden} and LSTCL \cite{wang2022long} embody a similar idea of learning to encode long-term temporal cues in clip-level representations by maximizing the similarity between a pair of short and long clips from the same video. The \emph{multiscale} aspect of \modelabb distinguishes it from BraVe and LSTCL. While these methods help the video model to extrapolate the contextual information contained in the longer clip from the short clip, their learning objective does not explicitly encourage it to understand how the contextual information may change over different durations. This may limit the video model's ability to understand the relations between short actions that occur within a few frames and long-term actions that may span several seconds or more. In contrast, by learning to predict future contextual information over varying temporal spans, \modelabb may enable the trained video model to gain a deeper understanding of actions at different levels of abstraction and recognize complex actions by identifying their sub-actions.


\noindent\textbf{Action forecasting.} State-of-the-art approaches \cite{damen2020rescaling,girdhar2021anticipative} are often aimed at addressing the short-term problem formulation where the goal is to anticipate the action that will occur in the next $\tau_a$ seconds using the context of an observed video sequence of $\tau_o$ seconds. Prior approaches have proposed to address this task by leveraging free additional information in the query videos either by aggregating past temporal context \cite{furnari2020rolling, sener2020temporal} or predicting representations of future frames and video clips \cite{vondrick2016anticipating, wu2020learning}. Gong \etal \cite{gong2022future} also leverage fully-attentional models to compute a more effective understanding of long-term temporal dynamics in the partially observed videos to generate more accurate predictions in the more challenging task of long-term forecasting \cite{damen2020rescaling,farha2020long,girdhar2021anticipative,grauman2022ego4d,sener2020temporal}. However, these strongly-supervised approaches often leverage pretrained visual representations that do not encode the multiscale nature of actions in videos, which limits their effectiveness. As such, \modelabb is orthogonal to these methods since we aim to learn more efficient base representations for downstream long-term forecasting tasks. We leave it to future work to integrate multiscale representations into state-of-the-art forecasting approaches.
\section{\modelname}

Our goal is to learn robust video representations that generalize well to downstream long-term forecasting tasks from a set of unlabeled videos. To this end, we introduce a self-supervised \modelname (\modelabb) objective, that aims to enable a video model to generate more accurate fine-grained action predictions of the forthcoming video clips given context from a partially observed clip sequence. Our approach is motivated by the reasoning that long-term forecasting requires the key capability of predicting the occurrences of future events at multiple timescales (\eg near and distant future). Similarly, \modelabb requires a video model to infer the initial context of the video from an observed clip sequence and leverage the context to condition its predictions of information that is contained in future clips. Due to a lack of explicit annotations during pretraining, we propose to exploit the \emph{multiscale} nature of complex actions in long videos for pseudo supervision. For example, the complex action of making an omelette can be decomposed into shorter atomic actions including cutting the onions and cracking the eggs. More specifically, \modelabb trains the video model to predict fine-grained spatiotemporal representations of the future that have been contextualized by aggregating information over varying numbers of future clips. We hypothesize that this objective encourages a video model to learn representations that encode future contextual information over multiple temporal spans. 

Unlike state-of-the-art video pretraining approaches \cite{feichtenhofer2021large,oord2018representation,qian2021spatiotemporal, wang2022long} which generally encourage different clips of the same video to have similar representations, MVP trains a video model to effectively represent the spatial and temporal structure of the observed video to extrapolate long-term information about future short and long actions. Intuitively, understanding the hierarchy of actions enables the video model to better reason about and recognize complex actions by identifying their sub-actions. Such an understanding may help the model to compute a more accurate prior distribution to condition its predictions on.

\subsection{Temporal aggregation of video clip sequences} \label{sec:video_model}
 While state-of-the-art video pretraining methods \cite{feichtenhofer2021large,qian2021spatiotemporal} often utilize pairs of video clips from the same video, our \modelabb objective trains a video model with pairs of video clip sequences $V^O$ and $V^F$ instead. \modelabb requires the video model to observe $V^O$ and infer the knowledge required to predict future contextual information that have been aggregated over the clips in $V^F$ at multiple timescales. To begin, we partition an input video into non-overlapping clips of 8 frames each (about 0.8s) and randomly sample the observed as well as future clip sequences $V^O = \{V_1^O, \cdot \cdot \cdot, V_{N_O}^O\}$ and $V^F = \{V_{N_O + K}^O, \cdot \cdot \cdot, V_{N_O + K + N_F}^F\}$, where $N_{O}$, $N_F$, and $K$ denote the number of observed, future, and temporal offset clips, respectively. We also define the temporal stride $S$ as the difference in number of clips between two timescales. Thus, \modelabb makes $N_P$ predictions, where $N_P = \frac{N_F}{S}$.

\begin{figure}[t]
\hspace{-14pt}
    \centering
    \includegraphics[width=0.5\textwidth]{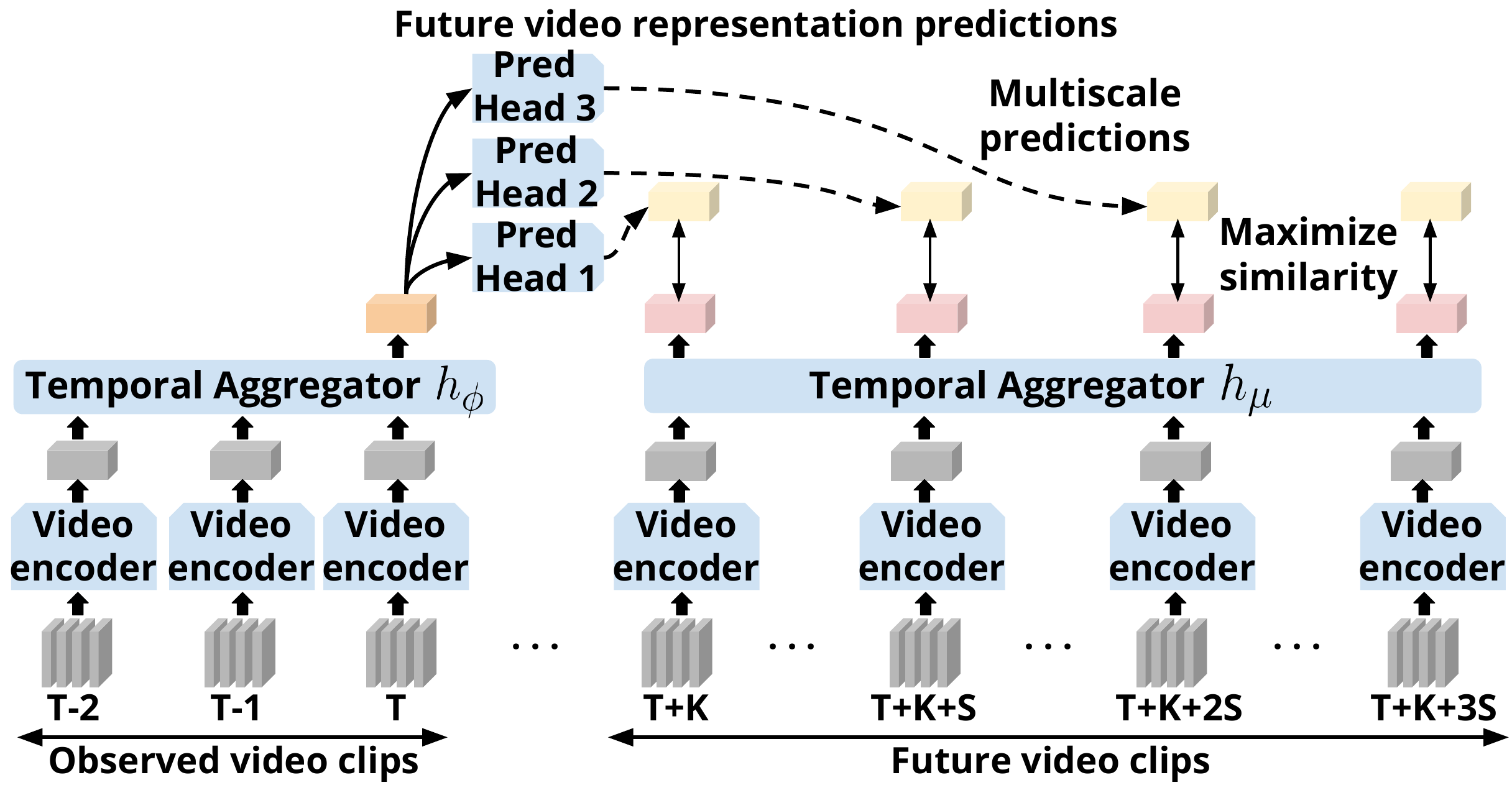}
    \caption{\textbf{\modelname.} Given an observed sequence of video clips, \modelabb learns to extract information that is required to predict contextualized representations of future video clips over multiple timescales.} 
\label{fig:model_fig}
    \vspace{-15pt}
\end{figure}

 Our video model (Figure~\ref{fig:model_fig}) is comprised of a video clip encoding function $g_\theta$ as well as temporal context aggregation functions $h_{\phi}$ and $h_{\mu}$. $g_\theta$ is used to encode an input clip into a set of spatiotemporal region representations while $h_{\phi}$ and $h_{\mu}$ are used to aggregate the temporal context of the observed and future clip sequences, respectively, by combining information over their constituent clip representations.


Due to the computationally demanding nature of our \modelabb objective, we adopt the lightweight yet powerful Multiscale Vision Transformers (MViT) \cite{fan2021multiscale} as our base encoder $g_\theta$ without modifications, which has been shown to outperform prior video encoders in action recognition despite containing significantly fewer parameters. We encode the $i$-th video clip as: $f_i = g_\theta(V_i), f_i \in \mathbb{R}^{L \times H \times W \times D}$ where $L$, $H$, $W$ and $D$ denote the temporal, height, width and channel dimensions, respectively. Then, we compute contextualized representations for both input sequences by aggregating information over the clips:
\begin{equation}
 z^O = z^O_{N_O} = h_{\phi}(g_{\theta}(V^O)), \enspace z^F = z^F_{N_F} = h_{\mu}(g_{\theta}(V^F)), 
\end{equation}
where $z^O$ and $z^F$ are the contextualized representations for the observed and future sequences, respectively.

\subsection{Spatiotemporal multi-head self-attention}
We argue that learning to predict fine-grained region representations over the spatial and temporal dimensions may be beneficial to understanding interactions between objects and actions in videos, unlike prior work focused on predicting global clip representations \cite{oord2018representation, qian2021spatiotemporal, wang2022long}. To this end, we train our model to predict spatiotemporal region representations of future clip sequences that have been contextualized over multiple timescales. This requires our temporal aggregation function to be able to compute contextual information between different spatial regions across multiple time steps. Intuitively, this objective can only be achieved with a strong understanding of the movement of objects over time and their relevance to different actions. 

A widely adopted convention for learning this function is to use multi-head self-attention (MHA) \cite{bertasius2021space} over the entire set of spatiotemporal regions in the video clip sequence. However, since self-attention has quadratic complexity, the computational requirements increase rapidly even for short sequences of video clips. To address this, we only aggregate temporal context information between video clips by computing self-attention over all regions at the same spatial locations in the video clip sequence. This is motivated by our observation that the output region representations of MViT for each time step have already been contextualized by aggregating information over other spatial regions, since the self-attention operation is an implicit function composited in the final video clip encoding function learnt by the MViT model. We refer interested readers to \cite{fan2021multiscale} for more details on the MViT architecture.

To begin, given an input spatiotemporal block $S \in \mathbb{R}^{L \times H \times W \times D}$, we project the set of temporal region features for the $j$-th spatial location $S_j \in \mathbb{R}^{L \times D}$, where $j \in hw$,  into its queries, keys and values:
\vspace{-5pt}
\begin{equation}
 S_{j,q} = S_j W_q, 
   \quad S_{j,k} = S_j W_k, 
\quad S_{j,v} = S_j W_v,
\vspace{-10pt}
\end{equation}
where $W_q$, $W_k$ and $W_v$ are the query, key and value projection weights of dimensions $D \text{x} D$. Then, we compute contextualized representations for the sequence using the MHA operation as follows:
\vspace{-10pt}
\begin{equation}
\hspace{-5pt}
    \text{MHA}(S_{j,q}, S_{j,k}, S_{j,v}w) = S_{j,v} \text{  Softmax}\left(\frac{S_{j,q} ^T S_{j,k}}{\sqrt{D}}\right)
    \vspace{-10pt}
\end{equation}
For a given spatiotemporal region representation $z_{i,t,h,w}$ from the i-th video clip, we compute its contextualized representations as: $z_{i,t,h,w}^{'} = \text{MHA}(z_{i,t,h,w})$. Finally, we predict the $j$-th future region representation at the $k$-th time step with a temporal stride of $S$ by passing the contextualized spatiotemporal region representations through a two-layer multilayer perceptron (MLP), \ie, $\hat{z}_{i,t,h,w} = \text{MLP}_k(z_{i,t,h,w}^{'})$. The entire set of predicted region representations $\hat{z}$ is used in Section~\ref{sec:multiscale_loss} to compute the training loss. Note that we use a different prediction head for each predicted timestep.


\begin{table*}[t]
{
\setlength{\tabcolsep}{3pt}
\begin{center}
\begin{tabular}{|c|c|c|ccc|ccc|ccc|}
\hline
Pretraining & Multiple & Pretraining  &  & Ego4D $\uparrow$ &  &  & EK55 $\uparrow$ & &  & EK100 $\uparrow$ & \\
\cline{4-12}
approach & clips used & supervision & Verb & Noun & Mean & Verb & Noun & Mean & Verb & Noun & Mean\\
\hline
\rowcolor{lightgray} Action recognition & No & Strong & 20.70 & 14.41 & 17.56 & 18.11 & 11.48 & 14.80 & 18.82 & 12.46 & 15.64\\
CVRL \cite{qian2021spatiotemporal} & No & Self & 25.90 & 25.85 & 25.88 & 22.17 & 17.07 & 19.62 & 22.92 & 16.60 & 19.76\\
CPC \cite{oord2018representation} & Yes & Self & 27.26 & 26.57 & 26.91 & 23.00 & 17.24 & 20.13 & 23.16 & 17.06 & 20.11\\
LSTCL \cite{wang2022long} & Yes & Self & 26.82 & 27.76 & 27.29 & 23.59 & 18.52 & 21.05 & 23.47 & 17.15 & 20.31\\
DPC \cite{han2019video} & Yes & Self & 28.18 & 29.03 & 28.61 & 24.02 & 19.03 & 21.52 & 25.25 & 18.18 & 21.72\\
CVRL \cite{qian2021spatiotemporal} & Yes & Self & 28.27 & 29.74 & 29.00 & 23.91 & 18.32 & 21.12 & 24.94 & 19.24 & 22.09\\
CONSTCL \cite{yuan2022contextualized} & Yes & Self & 27.49 & 29.13 & 28.31 & 24.47 & 19.52 & 22.00 & 25.41 & 19.35 & 22.38\\
\modelabb (Ours) & Yes & Self & \textbf{30.18} & \textbf{32.33} & \textbf{31.25} & \textbf{25.83} & \textbf{20.78} & \textbf{23.31} & \textbf{26.69} & \textbf{20.18} & \textbf{23.44}\\
\hline 
\end{tabular}
\end{center}
\vspace{-10pt}
\caption{\textbf{Order-agnostic long-term forecasting.} We report the mean average precision over all verb and noun classes. We see that self-supervised pretraining is generally more beneficial for long-term forecasting tasks than action recognition.}
\vspace{-10pt}
\label{tab:order_unaware_forecasting_results}
}
\end{table*}

\subsection{Multiscale targets and loss formulation}\label{sec:multiscale_loss}
To compute the prediction targets for self-supervision, we apply the aggregation function $h_\mu$ to $V_F$ in a causal manner, \ie the set of contextualized spatial region representations $S_{t,j}$ for the $j$-th spatial region at the $l$-th time step is computed by attending only to the regions that precede it temporally. For the $b$-th sequence of future video clips in a sampled batch, we extract a set of target representations $Z_b = \{z_{b,k}\}$, where $k \enspace \% \enspace S = 0$ and $Z_b \in \mathbb{R}^{N_P \times LHW \times D}$. Given a batch of unlabeled videos, we train the video model end-to-end using a contrastive loss \cite{oord2018representation} formulation as:
\begin{equation} 
A = \sum^B_{b=1} \sum^{N_P}_{j=1} \sum^{LHW}_{n=1} -\log\frac{\exp(\hat{z}_{b,j,n} \cdot z_{b,j,n} / \tau)}{%
\splitfrac{\textstyle
            \exp(\hat{z}_{b,j,n} \cdot z_{b,j,n} / \tau) +} 
          {\hspace*{-25pt}\textstyle 
           \sum\limits_{(b',j', n') != (b,j,n)}\exp(\hat{z}_{b,j,n} \cdot z_{b', j', n'} / \tau)}},
\end{equation}
where $\tau$ denotes the temperature value.
\section{Experiments}

\subsection{Downstream tasks}
We compare our \modelname objective to state-of-the-art self-supervised video pretraining methods on the tasks of \emph{order-agnostic} and \emph{specific} long-term action forecasting as well as video summary forecasting. We pretrain the video models on Ego4D \cite{grauman2022ego4d} and finetune them on both Ego4D and EpicsKitchen-55/100 \cite{Damen2018EPICKITCHENS,damen2020rescaling} for the downstream tasks. Additionally, we use a transformer encoder \cite{vaswani2017attention} and the meanpooling operation as our temporal context aggregators $h_\phi$ and $h_\mu$ (Section~\ref{sec:video_model}), respectively. We refer readers to the supplemental for more details of these datasets, implementation and baseline models.


\noindent\textbf{Order-agnostic action forecasting.} In order-agnostic long-term forecasting, we observe K\% of a video of duration T and predict if an action will occur within the remaining video. Given a vocabulary of $N_{\text{verb}}$ and $N_{\text{noun}}$ classes, we predict a $N_{\text{verb}}$-dimensional and $N_{\text{noun}}$-dimensional binary vectors, where each dimension indicate the probability of the class occurring in the future. We formulate this as a multi-label prediction task and finetune all pretrained models by optimizing the binary cross-entropy loss computed over all verb and noun classes. We compute the mean average precision (mAP) over all verb and noun classes.

\noindent\textbf{Order-specific action forecasting.} The order-specific task is a much more challenging setting, where the model is penalized even if it predicts the correct verb or noun but in the wrong order. Since the accuracy of the predicted actions depends on their temporal ordering, this can be formulated as a sequence prediction task. We finetune the pretrained models by optimizing the total cross-entropy losses for both verbs and nouns computed over all time steps. We adopt the edit distance metric \cite{grauman2022ego4d} to quantify how dissimilar the predicted and ground-truth action  sequences are to each other.

\noindent\textbf{Video summary forecasting.} In this multimodal task, for a video $V$ of $T$ temporal clips and an observed subsequence of length $T^O$, the goal is to retrieve its corresponding summary from a set of distractor summaries. Given the video $V$ and its summary $L$ containing $N_L$ words, we first extract the contextualized representation for the observed clip sequence: $c_{T^O} = h_\theta^{\text{agg}} (g_\theta^V (V_{0:T^O}))$. We extract a natural language representation $f_L \in \mathbb{R}^{L \text{x} D_L}$ for the summary using the pretrained BERT-Base \cite{devlin2018bert} model: $f_L = k_\phi(L)$, where $D_L$ is the output dimension of the BERT model and $k_\phi$ denotes the BERT model that is parameterized by $\phi$. We use linear layers $W_V$ and $W_L$ to project the video and language representations into the joint visual-semantic embedding space and finetune the models by optimizing the following contrastive loss formulation:
\begin{equation}
    L = \sum^B_{b=1} -\log \frac{\exp(c_{b,T^O} \cdot f_{b,L} / \tau)}{\splitfrac{\textstyle
            \exp(c_{b,T^O} \cdot f_{b,L} / \tau) +}
          {\textstyle 
           \sum\limits_{m \neq b}\exp(c_{b,T^O} \cdot f_{m,L} / \tau)}}.
\end{equation}
Intuitively, this objective encourages the model to learn an alignment between the video and language representations by maximizing the similarity between corresponding pairs of videos and text summaries. Consistent with prior work in text-to-video retrieval \cite{zhou2018towards}, we adopt the Recall@$K$ metric which computes the percentage of times the ground-truth summary is ranked in the top $K$ retrievals.

\begin{table*}[t]
{
\setlength{\tabcolsep}{3pt}
\begin{center}
\begin{tabular}{|c|c|c|ccc|ccc|ccc|}
\hline
Pretraining & Multiple & Pretraining &  & Ego4D $\downarrow$ &  &  & EK55 $\downarrow$ & &  & EK100 $\downarrow$ & \\
\cline{4-12}
approach & clips used & approach & Verb & Noun & Action & Verb & Noun & Action & Verb & Noun & Action\\
\hline
\rowcolor{lightgray} Action recognition & No & Strong & 0.754 & 0.901 & 0.977 & 0.741 & 0.947 & 0.962 & 0.758 & 0.952 & 0.969\\
CVRL \cite{qian2021spatiotemporal} & No & Self & 0.746 & 0.845 & 0.960 & 0.719 & 0.926 & 0.948 & 0.753 & 0.948 & 0.954\\
CPC \cite{oord2018representation} & Yes & Self & 0.735 & 0.838 & 0.956 & 0.719 & 0.936 & 0.951 & 0.746 & 0.944 & 0.954\\
LSTCL \cite{wang2022long} & Yes & Self & 0.752 & 0.846 & 0.963 & 0.721 & 0.935 & 0.950 & 0.739 & 0.939 & 0.950\\
DPC \cite{han2019video} & Yes & Self & 0.734 & 0.821 & 0.950 & 0.708 & 0.927 & 0.946 & 0.738 & 0.932 & 0.951\\
CVRL \cite{qian2021spatiotemporal} & Yes & Self & 0.735 & 0.822 & 0.952 & 0.719 & 0.926 & 0.948  & 0.735 & 0.930 & 0.948\\
CONSTCL \cite{yuan2022contextualized} & Yes & Self & 0.735 & 0.818 & 0.951 & 0.704 & 0.922 & 0.946 & 0.732 & 0.930 & 0.948\\
\modelabb (Ours) & Yes & Self & \textbf{0.724} & \textbf{0.809} & \textbf{0.943} & \textbf{0.690} & \textbf{0.908} & \textbf{0.941} & \textbf{0.721} & \textbf{0.918} & \textbf{0.942}\\
\hline 
\end{tabular}
\end{center}
\vspace{-10pt}
\caption{\textbf{Order-specific long-term forecasting evaluation.} We use edit distance as the metric and report performance on verb, noun and action classes. An action class is a combination of its verb and noun classes. The results suggest that learning to understand the multiscale nature of videos is crucial for making accurate fine-grained predictions.}
\label{ego4d_lta_results}
\vspace{-10pt}
}
\end{table*}

\subsection{Quantitative results}
\subsubsection{Order-agnostic long-term forecasting}
\label{subsec:order-unaware}
We aim to evaluate the effectiveness of our proposed \modelabb pretraining approach at learning video representations that encode future context over different temporal horizons. As such, we predict the future actions over the next 8 time steps and report the results on Ego4D, EK55 and EK100 in Table~\ref{tab:order_unaware_forecasting_results}. We observe that self-supervised video pretraining is generally more beneficial to tasks requiring the key capability of long-term forecasting as compared to the strongly supervised variant of action recognition (first row of Table~\ref{tab:order_unaware_forecasting_results}). Despite not requiring human-annotated labels during pretraining, our proposed \modelabb approach leads to approximately 14\% improvement in future verb and noun predictions over its strongly-supervised counterpart when finetuned on the Ego4D task annotations. We hypothesize that the learning objective of predicting future clip representations is crucial for action anticipation.

We also observe across all datasets that the state-of-the-art pretraining objective of learning clip-invariant video representations \cite{qian2021spatiotemporal,feichtenhofer2021large} does not generalize well to downstream tasks that require effective reasoning over clip sequences. In fact, simply extending the aforementioned pretraining objective to maximize the similarity between representations of two clip sequences sampled from the same video leads to significant improvements in future action predictions, especially over the longer temporal horizon of 8 clips. Our \modelabb approach also outperforms LSTCL \cite{wang2022long} by a significant margin (\eg, we obtain a 3-5\% improvement on Ego4D). Since LSTCL aims to encode long-term temporal cues in video representations of shorter clip sequences, our gains suggest that learning to predict contextual information of future clip sequences serves as an effective pretraining objective for long-term video understanding.

\subsubsection{Order-specific long-term forecasting} 
Table~\ref{ego4d_lta_results} reports the results across all three datasets  on the more challenging task of predicting actions at specific time steps. Similar to our results for the order-unaware task in Section~\ref{subsec:order-unaware}, we also observe that our proposed \modelabb approach generalizes better to a task that requires accurate fine-grained predictions. We note that pretraining approaches that learn to predict future clip representations at the fine-grained region-level such as DPC, CONSTCL and ours generally perform better under this challenging setting as compared to variants that predict global representations of future video clips including CPC and CVRL. One possible reason is that predicting fine-grained spatiotemporal region representations in the future is a much more challenging objective that necessitates the video model to understand the structure of different atomic actions in untrimmed videos. In particular, our gains across all three datasets suggest that learning to predict future region-level representations is especially crucial for verb predictions. This is evidenced by the much larger margins of improvement achieved by such approaches in predicting verbs in future clips as compared to nouns. For example, \modelabb reduces the edit distances by 0.029 and 0.018 on verb and noun predictions, respectively. In contrast to the order-agnostic task, we see that the improvements achieved by our \modelabb objective are smaller, which further emphasizes the difficulty of predicting actions precisely at specific timesteps. 

Additionally, we aim to understand the effectiveness of learning to predict future contextual information that is aggregated from video clips over different temporal horizons. In particular, we compare against CONSTCL \cite{yuan2022contextualized}, which also aims to reconstruct fine-grained spatiotemporal region representations of a future video clip sequence given the context of an observed clip sequence. Despite not relying on pretrained object detectors to identify location priors, our proposed \modelabb approach outperforms CONSTCL on both verb and noun predictions (\eg reducing edit distance by 0.008 on Ego4D) while only using dense spatiotemporal feature maps. We hypothesize that our pretraining objective of predicting aggregated future spatiotemporal region representations helps a video model to better reason about the correlations between different atomic actions and how they contribute to the overarching goal in videos.


\subsubsection{Video summary forecasting}
Finally, Table~\ref{ego4d_summary_prediction_results} reports our results on the multimodal video summary forecasting task. Besides video-only tasks, we note that the self-supervised pretraining approaches also generalize much better to a downstream task that involves the language modality than the strongly-supervised task of action recognition. Unlike the results on the previous tasks of order-unaware and specific long-term forecasting, we observe that the pretraining objective of learning clip-invariant video representations such as CVRL (single and multiple clips) and LSTCL outperforms DPC by a substantial margin of $1 - 5\%$ in R@1 accuracy. 

We hypothesize that this may be due to the DPC pretraining approach training the video model to predict the representations of consecutive video clips in the future. In contrast, the aforementioned approaches sample the observed and predicted video clip sequences from the same video but at randomly determined times. This may encourage the video model to learn to extrapolate the contextual information further into the future instead of always predicting the \emph{immediate} future as in the case of the DPC method. Interestingly, we also observe that learning to predict fine-grained spatiotemporal region representations during pretraining may be not be as critical for understanding the overarching context of a video as the previous evaluation tasks. This is evidenced by the fact that CVRL pretrained with multiple video clips actually outperforms CONSTCL by $4\sim\%$ in R@1 accuracy. Lastly, the performance gains of approximately $3 - 8\%$ in R@1 accuracy achieved by our proposed \modelabb approach over CVRL clip sequence, LSTCL and CONSTCL suggest that learning to reason about aggregated future contextual information over multiple time scales is especially beneficial to helping a model to extrapolate the semantics of the entire video.

\begin{table}[t!]
{
\setlength{\tabcolsep}{2pt}
\begin{center}
\begin{tabular}{|c|c|c|ccc|}
\hline
\small{Pretraining} & \small{Multiple} & \small{Pretraining} &  &  & \\
\small{approach} & \small{clips} & \small{supervision} & \small{R@1}$\uparrow$ & \small{R@5}$\uparrow$ & \small{R@10}$\uparrow$ \\
\hline
\rowcolor{lightgray} \small{Action recognition} & \small{No} & \small{Strong} & 0.90 & 5.00 & 8.80\\
\small{CPC \cite{oord2018representation}} & \small{Yes} & \small{Self} & 9.70 & 28.60 & 41.80 \\
\small{DPC \cite{han2019video}} & \small{Yes} & \small{Self} & 10.10 & 29.70 & 43.20 \\
\small{CVRL \cite{qian2021spatiotemporal}} & \small{No} & \small{Self} & 11.00 & 34.80 & 49.50 \\
\small{LSTCL \cite{wang2022long}} & \small{Yes} & \small{Self} & 12.70 & 38.90 & 53.10\\
\small{CONSTCL \cite{yuan2022contextualized}} & \small{Yes} & \small{Self} & 11.40 & 41.80 & 53.90\\
\small{CVRL \cite{qian2021spatiotemporal}} & \small{Yes} & \small{Self} & 15.90 & 40.70 & 56.50\\
\small{\modelabb (Ours)} & \small{Yes} & \small{Self} & \textbf{19.30} & \textbf{50.70} & \textbf{65.00}\\
\hline 
\end{tabular}
\end{center}
\vspace{-10pt}
\caption{\textbf{Video summary forecasting on the Ego4D dataset. } MVP helps the video model to learn more robust representations that generalize better than prior work to the multimodal task of text summary retrieval.}
\vspace{-10pt}
\label{ego4d_summary_prediction_results}
}
\end{table}



\subsubsection{Ablation studies}
We ablate different aspects of MVP approach to determine their relative contributions to the robustness of the learnt representations. Specifically, we compare the effectiveness of the representations of different model variants on the downstream task of order-unaware forecasting on Ego4D. 

\begin{figure}[h!]
    \centering
\includegraphics[width=0.5\textwidth]{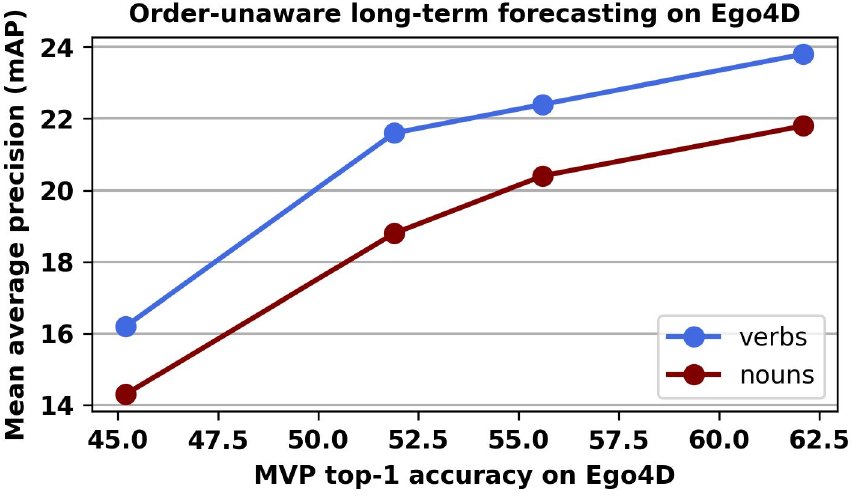}
    \vspace{-20pt}
    \caption{\textbf{Benefit of \modelabb.} We study the relation between self-supervised pretraining prediction accuracy and mean average precision on order-agnostic long-term forecasting.}
    \label{fig:pretraining_acc_plot}
\end{figure}

\begin{table}[h]
\begin{center}
\begin{tabular}{|c|ccc|ccc|}
\hline
Temporal offset $K$ & Verb $\uparrow$ & Noun $\uparrow$ & Mean $\uparrow$\\
\hline
1 & 23.47 & 21.10 & 22.28 \\
4 & 27.15 & 26.09 & 26.62 \\
8 & 27.95 & 26.78 & 27.37 \\
12 & 26.39 & 25.98 & 26.18 \\
16 & 27.88 & 26.09 & 26.99 \\
Geometric & 26.80 & 25.99 & 26.39 \\
Random (ours) & \textbf{30.18} & \textbf{32.33} & \textbf{31.25} \\
\hline 
\end{tabular}
\end{center}
\vspace{-10pt}
\caption{\textbf{Temporal offset ablation on Ego4D.} We ablate the effect of the temporal offset during pretraining on the downstream task of order-unaware long-term forecasting.}
\label{ego4d_order_unaware_temporal_offset_ablation}
\vspace{-15pt}
\end{table}

\begin{figure*}[t] 
    \centering
\includegraphics[width=\textwidth]{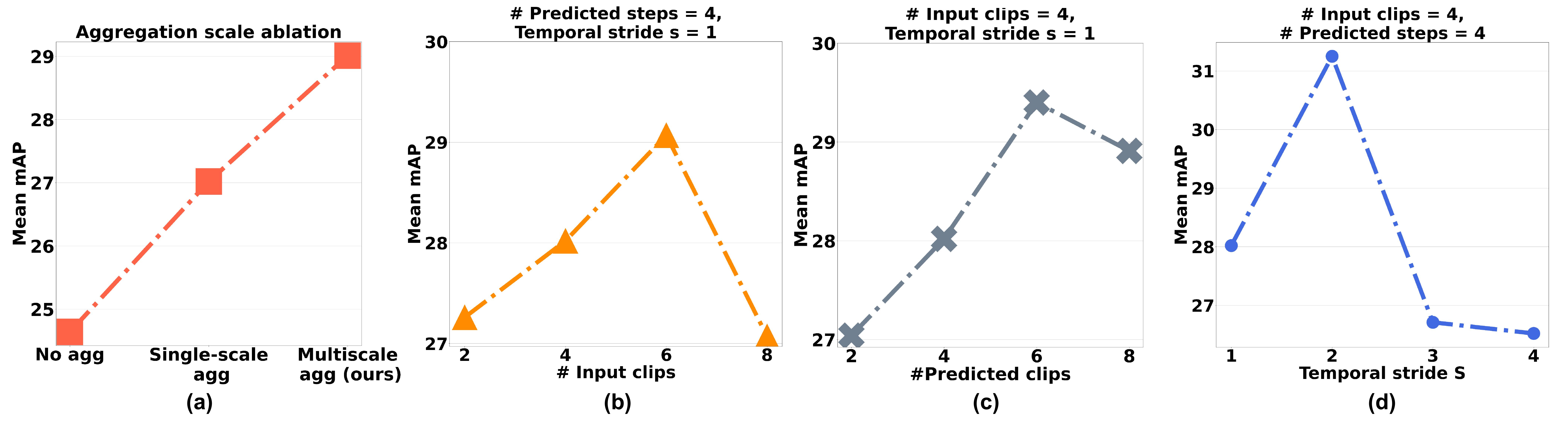}
    \vspace{-20pt}
    \caption{\textbf{Ablation of \modelabb.} (a) The results suggest that learning to model the temporal dynamics in videos at multiple timescales is crucial for action forecasting. (b) Providing more context with more observed video clips is generally helpful for learning more robust representations. (c) Increasing the number of predicted steps helps the video model to make more accurate action predictions to a certain degree. (d) Using a small temporal stride to aggregate context in the future clip sequence over multiple timescales is more beneficial than higher values.}
    \label{fig:ablations_plot}
    \vspace{-10pt}
\end{figure*}

\noindent\textbf{Effectiveness of \modelabb.} We evaluate the benefit of our \modelname approach in Figure~\ref{fig:pretraining_acc_plot} by studying the correlation between the prediction accuracy of the video model during pretraining and the downstream performance by using checkpoints at various stages of pretraining. While \modelabb uses a contrastive formulation, we compute the prediction accuracy as the percentage of predicted regions that have the highest similarity with their ground-truth counterparts. We observe a direct correlation between the prediction accuracy during pretraining and the mean mAP score over all verb and noun classes, which suggests that learning to encode the multiscale nature of videos in the base representations is beneficial for long-term forecasting tasks.

\noindent\textbf{Temporal offset $\mathrm{\mathbf{K}}$.} In Table~\ref{ego4d_order_unaware_temporal_offset_ablation}, we observe that verb and noun prediction accuracy increases as we increase $K$ during pretraining. This is unsurprising since the video model should be able to better predict future actions by learning to reason about the contextual information further into the future during pretraining. However, we also see that using a temporal offset of 12 clips actually leads to a drop in performance. One possible reason is that the future is non-deterministic and predicting information too far into the future introduces a high degree of noise during pretraining. We also hypothesize that sampling random temporal offset values works the best because learning to predict future contextual information over varying temporal horizons acts as a form of regularization and prevents the model from overfitting to predictions over a constant temporal period.

\noindent\textbf{Multiscale benefits.} We investigate the importance of multiscale aggregation during pretraining on downstream performance (Fig~\ref{fig:ablations_plot}(a)). Specifically, we train the video model with a variant of \modelabb where we only predict the uncontextualized representations of future clips (no aggregation) and another where the aggregation of context is computed over a single scale. To begin, we observe the importance of predicting contextualized representations, where predicting uncontextualized clip representations results in a drop of $2\sim$ \% in mean mAP. More importantly, we also see that learning to predict future clip representations that are aggregated over multiple timescales results in a significant improvement over predicting those that are only contextualized over a single timescale. These results may support our hypothesis that learning to understand the multiscale nature of actions helps the video model to better infer the underlying goals and thus, anticipate future actions.


\noindent\textbf{Number of input clips $\mathrm{\mathbf{N_O}}$.} In Figure~\ref{fig:ablations_plot}(b), we observe that increasing the number of clips in the observed sequence $V^O$ during pretraining generally leads to better downstream performance. However, we see that the forecasting results drop when we use 8 input clips. One possible reason is that using more input clips results in more observed context which may ease the difficulty of the pretraining objective and consequently, reducing the robustness of the learnt representations to downstream forecasting tasks. 

\noindent\textbf{Number of predicted clips $\mathrm{\mathbf{N_P}}$.} We also aim to understand the importance of varying the number of predicted clips during pretraining on downstream forecasting performance in Figure~\ref{fig:ablations_plot}(c). Intuitively, setting a higher number of predicted future clips increases the difficulty of our \modelabb objective since the video has to learn to predict contextual information that is further out into the future. While increasing the number of predicted clips is generally beneficial for downstream performance, we also see that predicting 8 future clips results in a drop in performance. We theorize that it may be too hard to predict the contextualized information too far out into the future since it is non-deterministic. This may introduce some noise during pretraining which adversely affects the learnt video representations. 

\noindent\textbf{Temporal stride $\mathrm{\mathbf{S}}$ for aggregation.} Last but not least, we ablate the effect of the temporal stride $S$ during pretraining in Figure~\ref{fig:ablations_plot}(d). We obtain the best downstream performance when we increase the temporal stride from $1$ to $2$, which may suggest that a higher temporal stride encourages the video model to learn to encode longer-term future contextual information. We hypothesize that larger strides actually results in a significant drop in performance because it may be too challenging for the video model to learn to understand the structure and relationships between different atomic actions if they are very distant in time.

\subsection{Limitations} 
The target representations in \modelabb are computed by aggregating information over future clips using a fixed temporal stride for different timescales. However, this may not always be realistic since different complex actions can consist of varying numbers of atomic actions. 
\section{Conclusion}
In summary, we introduce \modelname, a self-supervised approach that aims to learn robust video representations for downstream long-term forecasting tasks. Given an observed video clip sequence, we train a video model to predict aggregated representations of future clips over multiple timescales. We demonstrate empirically that learning to encode future contextual information helps the video model to generalize better to long-term forecasting tasks than prior work, which highlights the importance of multiscale pretraining to long-term video understanding. Last but not least, we extract key insights on different aspects of \modelabb, through an extensive ablation study, that we hope will be beneficial to further research on learning multiscale video representations. Some interesting avenues for future work may include further exploring the capabilities of these representations for other video and multimodal tasks such as action recognition and text-to-video retrieval.

\noindent 
\textbf{Acknowledgements}: This material is based upon work supported, in part, by DARPA under agreement number HR00112020054. We would like to thank Gideon Stocek and Nishanth Alapati for their assistance with setting up the compute infrastructure for the experiments.

{\small
\bibliographystyle{ieee_fullname}
\bibliography{egbib}
}

\clearpage
\appendix

In this supplemental, we provide the following additional material to the main submission:
\begin{enumerate}
    \item[A.] Training and evaluation datasets details
    \item[B.] Implementation details
    \item[C.] Spatiotemporal constrastive loss formulation
    \item[D.] Baseline models for comparisons
\end{enumerate}

\section{Datasets}

\noindent\textbf{Ego4D}~\cite{grauman2022ego4d} is the largest dataset of egocentric videos spanning over 3600 hours of daily life activities ranging from household to outdoor leisure scenarios. These videos are collected by 931 camera-wearers from 9 different countries, who record their unscripted interactions as they engage in daily activities under a large variety of settings. In contrast to existing video recognition datasets, videos in Ego4D are generally much longer in duration since they span from 1 to 10 hours as compared to 10 seconds video clips in Kinetics 400/600 \cite{carreira2018short,carreira2019short}. Additionally, it is much larger in scale and diversity of activities than existing egocentric video datasets such as Epic-Kitchens 55/100 \cite{Damen2018EPICKITCHENS, damen2020rescaling}. Each video is also densely annotated by humans, who provide annotations describing notable interactions in the videos as well as high-level summaries. This dataset facilitates the exploration and further research in a variety of downstream tasks such as audio-visual diarization and forecasting. We use the provided annotations to evaluate our proposed MTPL approach on long-term forecasting as well as video summary predictions. We adopt the same splits for training and evaluation on the target tasks as Grauman \etal \cite{grauman2022ego4d}. In this dataset, we conduct our evaluations on the training and validation splits since the test evaluation is conducted on a held-out set via a submission to their challenge portal. We also note that the number of verb and noun classes present in all 3 provided splits are not consistent since each split contains some verb and noun classes that are not present in other splits. Please refer to the supplementary material for more details.
\smallskip

\noindent\textbf{EpicsKitchen-55/100.} EpicKitchens-100 (EK100) \cite{damen2020rescaling} is another large dataset of egocentric videos. Similar to Ego4D, it also provides 700 long unscripted egocentric videos that span approximately 100 hours. It is less diverse than Ego4D since the participants only engage in daily activities in the kitchen. EpicKitchens-55 (EK55) \cite{Damen2018EPICKITCHENS} is an earlier and smaller version of EK100 but it provides the same types of videos and annotations. We use EK55 and EK100 to evaluate on the tasks of order-agnostic and order-specific long-term forecasting. 

\begin{figure*}[t!]
    \centering
\includegraphics[width=\textwidth]{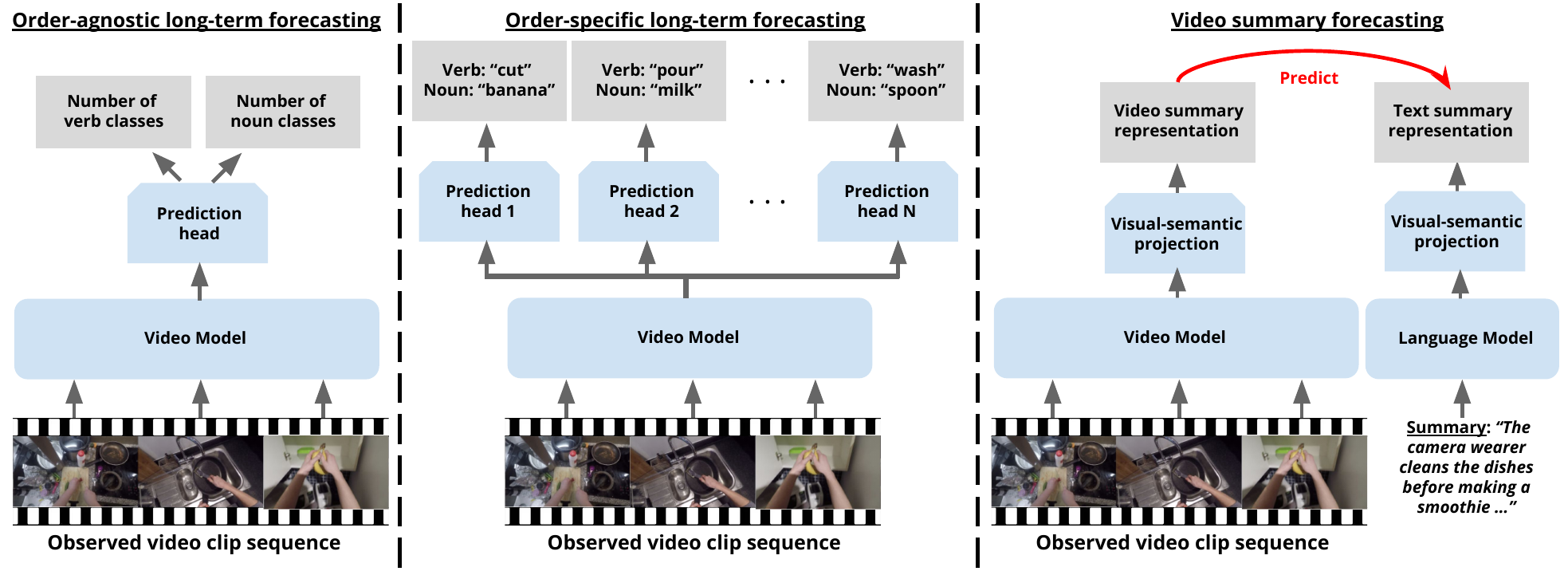}
    \caption{\textbf{Implementation for downstream long-term forecasting tasks.} We finetune our pretrained video models on the downstream tasks of order-agnostic and order-specific action forecasting as well as video summary forecasting on the target datasets with strong supervision.}
    \label{fig:downstream_tasks}
\end{figure*}

\section{Implementation details}

\subsection{\modelname}
We implement all models and experiments using the Pytorch deep learning library. We use the Multiscale Vision Transformer (MViT) \cite{fan2021multiscale} as our base video encoder and 1 transformer encoder layers with 1 attention heads as our temporal context aggregator. The MVIT encoder typically accepts a video clip of 16 frames as input and outputs a global clip representation, which is the contextualized output of the classification token. However, in our case, we reduce the number of frames per clip to 8 due to memory constraints. Additionally, we discard the classification token during pretraining and perform our future feature predictions at the spatiotemporal region granularity. During the second stage of finetuning, we compute a global clip representation by performing meanpooling over the spatiotemporal region representations. 

Since we sample the video frames at 10 frames per second (FPS), the temporal duration of each clip is approximately 0.8 seconds long. Each input video clip is preprocessed by randomly scaling the height of the frames between 248 and 280 pixels and taking crops of 224 x 224 pixels. During the first stage of pretraining on the Ego4D dataset, we also perform random augmentations to the video clips including random horizontal flipping and color jittering. The future feature prediction function is represented as a two-layer multilayer perceptron (MLP) with a non-linear ReLU operation and hidden dimension of 768.

\subsection{Downstream long-term forecasting tasks}
Figure~\ref{fig:downstream_tasks} illustrates how our pretrained video model and its learnt representations are transferred to the order-agnostic and order-specific action forecasting as well as video summary forecasting. To begin, given the sequence of $N_V$ observed video clips in each task $V = \{V_1, \cdot \cdot \cdot V_{N_V}\}$, we extract the contextualized representation of the last timestep as follows:
\begin{equation}
    z_{N_V} = h_\phi(g_\theta(Vz)), \quad z_{N_V} \in \mathbb{R}^D
\end{equation}
where $D$ is the output channel dimension. For all downstream tasks, we finetune linear probes on top of the pretrained video model, which is kept frozen.

\noindent\textbf{Order-agnostic action forecasting.} Given a vocabulary of $N_{\text{verb}}$ and $N_{\text{noun}}$ classes, we predict a $N_{\text{verb}}$-dimensional and $N_{\text{noun}}$-dimensional binary vectors as:

\begin{equation}
\begin{aligned}
    p_{\text{verb}} = f_{\text{verb}}(z_{N_V}), \\
    p_{\text{noun}} = f_{\text{noun}}(z_{N_V}),
\end{aligned}
\end{equation}
where each dimension in the predicted vectors indicates the probability of the verb or noun class occurring in the future. We formulate this as a multi-label prediction task and finetune all pretrained models by optimizing the binary cross-entropy loss computed over all verb and noun classes as:
\begin{equation}
    L = -\sum^B_{b=1}(\sum^{N_{\text{verb}}}_{i=1} y_{\text{verb},b,i} \log(p_{\text{verb},b,i}) + \sum^{N_{\text{noun}}}_{i=1} y_{\text{noun},b,i} \log(p_{\text{noun},b,i})),
\end{equation} where $y_{\text{verb},b,i}$ and $y_{\text{noun},b,i}$ are the ground-truth verb and noun binary labels, respectively.

\noindent\textbf{Order-specific action forecasting.} In this more challenging setting, the goal is to make fine-grained action predictions at specific timesteps. For simplicity, we adopt the same training and evaluation setup as in \cite{grauman2022ego4d} and use separate prediction heads for different timesteps. For each timestep, we formulate the subtask as a multiclass prediction problem for both verbs and nouns. Consequently, we finetune the pretrained video models using the following loss formulation:
\begin{equation}
    L = -\sum^B_{b=1} \sum^{N_P}_{t=1} (y_{\text{verb},b,t} \log(p_{\text{verb},b,t}) + 
    (y_{\text{noun},b,t} \log(p_{\text{noun},b,t})
    ).
\end{equation}

\noindent\textbf{Video summary forecasting.} As shown in Figure~\ref{fig:downstream_tasks} (right), we adopt the dual encoder architecture to address this multimodal task. Similar to prior work on learning joint visual-language representations including CLIP and ALIGN, we also use the late fusion mechanism where the semantic similarity between the final video and language representations are computed using a final dot product operation. 


\begin{figure}[h]
    \centering
    \includegraphics[width=0.5\textwidth]{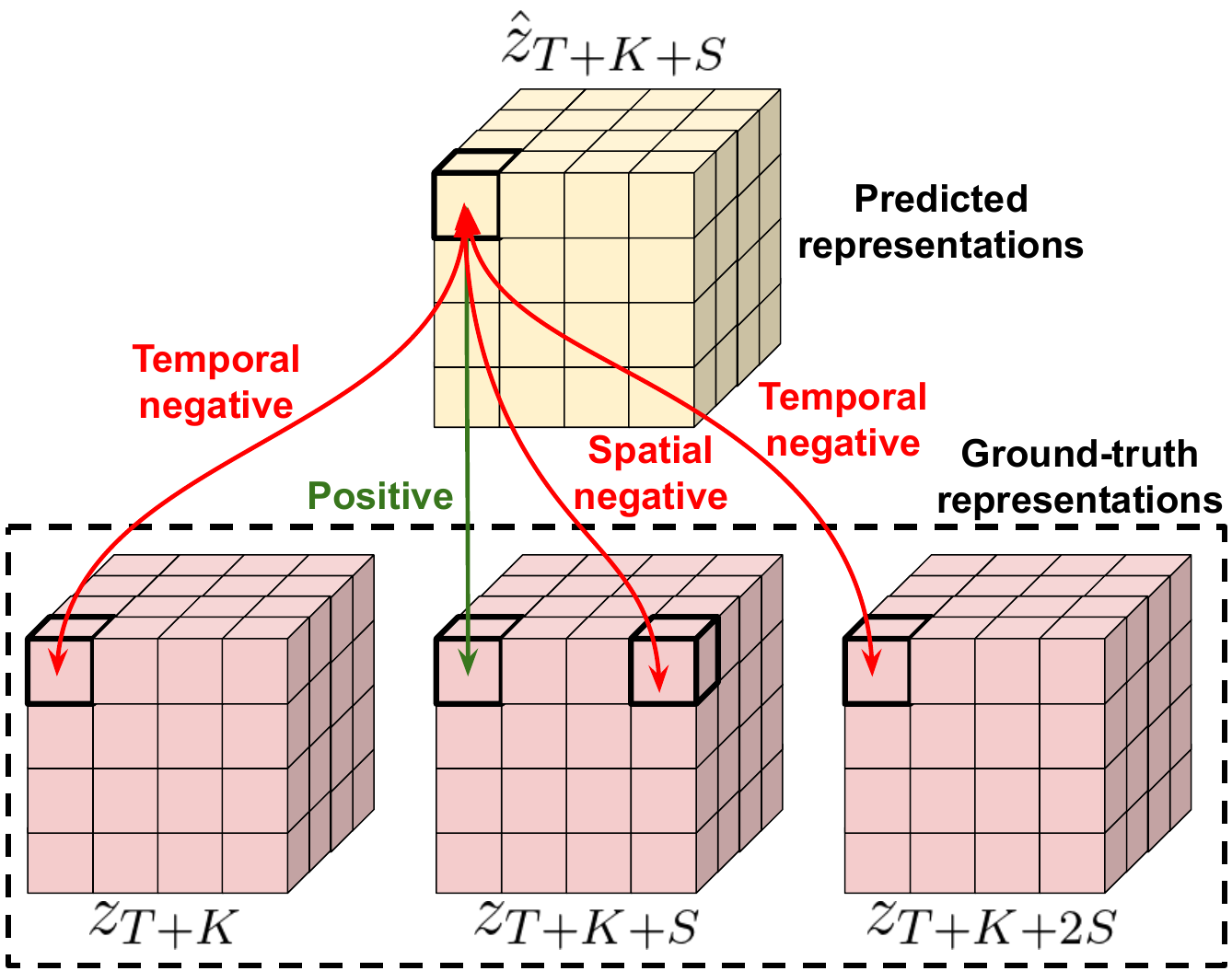}
    \caption{\textbf{Spatiotemporal region predictions.} Our \modelabb approach trains a video to predict future contextual information contained in fine-grained spatiotemporal regions.}
\label{fig:spatiotemporal_region_preds}
\end{figure}

\section{Spatiotemporal constrastive loss formulation}
We provide an illustration of how our proposed \modelabb objective trains a video model to predict fine-grained spatiotemporal region representations using the contrastive loss formulation in Figure~\ref{fig:spatiotemporal_region_preds}. Given the predicted representation of the $j$-th spatial region at the $t$-th timestep $\hat{z}_{t,j}$, we aim to maximize its semantic similarity with its ground-truth aggregated representation $z_{t,j}$ and the negative samples in the entire set of distractors consist of both hard negatives such as other spatial regions at the same timestep and easy negatives including representations that belong to clips from other videos in the sampled batch.
 
\section{Baseline models} We briefly describe the self-supervised video pretraining baselines that we compare our proposed \modelabb objective against in our evaluations.

\noindent\textbf{Contrastive predictive coding (CPC).} The Contrastive Predictive Coding (CPC) \cite{oord2018representation} approach aims to learn video representations that encode global information that is shared between different clips of a video. CPC uses the context from an observed clip sequence to predict the future \emph{uncontextualized} information in the future clips that directly follow after the observed sequence. It also uses multiple prediction heads for representations of different timesteps that it tries to predict for.

\noindent\textbf{Dense predictive coding (DPC).} The Dense Predictive Coding (DPC) \cite{han2019video} approach builds on top of CPC to learn video representations of predicting \emph{uncontextualized} information but conditions its predictions for a given timestep with the context of the predicted information at the preceding timestep. Additionally, unlike CPC, the DPC objective aims to compute spatiotemporal representations instead of global clip representations.

\noindent\textbf{Contrastive video representation learning (CVRL).} We also compare \modelabb to the Contrastive Video Representation Learning (CVRL)  \cite{qian2021spatiotemporal} approach, which is largely inspired by popular image-based self-supervised pretraining objectives \cite{chen2020simple,chen2020improved,chen2021exploring}. CVRL trains a video model to maximize the similarity between representations of different clips that are randomly sampled from the same videos. While we compare to CVRL in its vanilla setting which uses pairs of video clips, we also train and evaluate a variant of CVRL which maximizes the similarity between representations of pairs of clip sequences.

\noindent\textbf{Long-Short Term Contrastive Learning (LSTCL).} Similar to the CVRL approach, the Long-Short Term Contrastive Learning (LSTCL) \cite{wang2022long} is initially proposed to learn video representations by maximizing the similarity between representations of video clip pairs. During pretraining, it accepts as input a short clip and another long clip which contains temporal information that is not present in the former. LSTCL trains a video model to extrapolate past and future information from a small observed temporal window. We also extend LSTCL to train on pairs of video clip sequences with the same total number of video clips per sample during pretraining to facilitate fair comparisons.

\noindent\textbf{Contextualized Spatio-Temporal Contrastive Learning with Self-Supervision (CONSTCL).} Last but not least, we also compare to the Contextualized Spatio-Temporal Contrastive Learning with Self-Supervision (CONSTCL) \cite{yuan2022contextualized} approach. CONSTCL aims to address the limitation of spatiotemporal invariance \cite{feichtenhofer2021large} enforced by the CVRL objective. The CONSTCL objective leverages a region-based preraining task which trains the video model to transform video representations from one clip sequence to another, given the context from the first sequence.

\end{document}